\documentclass[runningheads]{llncs}
\usepackage{graphicx}
\usepackage{array}
\usepackage{subfigure}
\usepackage{setspace}
\usepackage{amssymb}
\usepackage{ulem}
\usepackage{booktabs}
\usepackage{amsmath}
\usepackage{xcolor}
\usepackage{multirow}
\usepackage{amsmath,amssymb,amsfonts}
\usepackage{textcomp}
\usepackage{algorithm}
\usepackage{algpseudocode} 
\usepackage[pagebackref=true,breaklinks=true,colorlinks=true,bookmarks=false,citecolor=blue,urlcolor=blue,linkcolor=blue]{hyperref}
\usepackage[misc]{ifsym}
\usepackage{bm}
\usepackage{fdsymbol}

\begin{document}
\title{Treatment Outcome Prediction for Intracerebral Hemorrhage via Generative Prognostic Model with Imaging and Tabular Data}

\titlerunning{Treatment outcome prediction for ICH}
\author{Wenao Ma\textsuperscript{1}, Cheng Chen\textsuperscript{2}, Jill Abrigo\textsuperscript{3}, Calvin Hoi-Kwan Mak\textsuperscript{4}, \\ Yuqi Gong\textsuperscript{1}, Nga Yan Chan\textsuperscript{3}, Chu Han\textsuperscript{5,6}, Zaiyi Liu\textsuperscript{5,6}, Qi Dou\textsuperscript{1(\Letter)}}

%1{Ma, Wenao}
%2{Chen, Cheng}
%3{Abrigo, Jill}
%4{Mak, Calvin Hoi-Kwan}
%5{Gong, Yuqi}
%6{Chan, Nga Yan}
%7{Han, Chu}
%8{Liu, Zaiyi}
%9{Dou, Qi}

% \author{Paper ID: 2676}
\institute{$^{1}$ Department of Computer Science and Engineering, The Chinese \\ University of Hong Kong, Hong Kong, China\\
$^{2}$ Center for Advanced Medical Computing and Analysis, \\ Harvard Medical School, Boston, USA \\
$^{3}$ Department of Imaging and Interventional Radiology, The Chinese \\ University of Hong Kong, Hong Kong, China \\
$^{4}$ Department of Neurosurgery, Queen Elizabeth Hospital, Hong Kong, China\\
$^{5}$ Department of Radiology, Guangdong Provincial People's Hospital (Guangdong Academy of Medical Sciences), Southern Medical University, Guangzhou, China \\
$^{6}$ Guangdong Provincial Key Laboratory of Artificial Intelligence \\ in Medical Image Analysis and Application, Guangzhou, China}
\authorrunning{Ma et al.}

\maketitle        
\begin{abstract}
Intracerebral hemorrhage (ICH) is the second most common and deadliest form of stroke. Despite medical advances, predicting treatment outcomes for ICH remains a challenge.
This paper proposes a novel prognostic model that utilizes both imaging and tabular data to predict treatment outcome for ICH. Our model is trained on observational data collected from non-randomized controlled trials, providing reliable predictions of treatment success. Specifically, we propose to employ a variational autoencoder model to generate a low-dimensional prognostic score, which can effectively address the selection bias resulting from the non-randomized controlled trials. Importantly, we develop a variational distributions combination module that combines the information from imaging data, non-imaging clinical data, and treatment assignment to accurately generate the prognostic score. We conducted extensive experiments on a real-world clinical dataset of intracerebral hemorrhage. Our proposed method demonstrates a substantial improvement in treatment outcome prediction compared to existing state-of-the-art approaches. Code is available at \href{https://github.com/med-air/TOP-GPM}{https://github.com/med-air/TOP-GPM}.
\end{abstract}

\section{Introduction}
Intracerebral Hemorrhage (ICH) is a bleeding into the brain parenchyma, which has the second-highest incidence of stroke (accounts for more than $10\%$ of strokes) and remains the deadliest type of stroke with mortality more than $40\%$ \cite{ducruet2009complement,feigin2009worldwide,flaherty2006long}. Timely and proper treatments are crucial in reducing mortality \cite{hemphill2015guidelines}, as well as improving functional outcomes, which is clinically deemed more valuable for prognostic model \cite{cheung2003use}. However, the treatment decision-making of ICH still remains problematic despite progression of clinical practice \cite{kim2017spontaneous}. 
% In clinical practice, the treatment decision making lacks unified standards and depends on multiple factors \cite{gregson2019surgical}. 
It is widely accepted that there is currently no effective approach in clinical practice to aid in decision-making regarding the evaluation of risks and benefits of a treatment \cite{godoy2006predicting,gregson2019surgical,steiner2014european}.
Thus, there is an urgent need for reliable treatment recommendation model in clinical practice. 
Unfortunately, existing works of ICH treatment outcome prediction can either predict the outcome under a certain type of treatment \cite{hall2021identifying,hemphill2001score,stein2010spontaneous}, or consider treatment assignment as an input variable but ignore potential differences in outcomes due to varying treatment assignments \cite{godoy2006predicting,guo2022machine,ji2013novel}, making it still challenging to determine from data which treatment would yield better outcomes. For this reason, we seek to provide a treatment recommendation model that outputs the reliable outcomes of all potential treatment assignments and focuses on the effect of different treatments.

One of the major challenges in treatment effect estimation is missing \textit{counterfactual} outcome \cite{johansson2016learning,peters2017elements}. This means that we can only observe the outcomes of the actual treatment decision made for an individual. As a consequence, the counterfactuals, that are, the outcomes that would have resulted from treatment decisions not given to the patient are missing. Another challenge is \textit{selection bias} brought by non-randomized controlled trials \cite{bica2021real}, that the treatment assignments may highly depend on patients' characteristics. For instance, for the ICH patients with a Glasgow Coma Scale (GCS) score 9–12 \cite{teasdale1974assessment}, early surgery is generally preferred over conservative treatment \cite{steiner2014european}. This selection bias can thus make the model unreliable in predicting the outcome of conservative treatment for patients with GCS 9-12 due to lack of observational data. These factors lead to inaccurate comparisons of treatment effects, as we can only get reliable outcome on one side (i.e., early surgery or conservative treatment).

To handle these challenges, some related works were based on the concept of balanced representation learning, which proposes to use additional loss to mitigate the aforementioned selection bias in the representation space \cite{johansson2016learning,shalit2017estimating,shi2019adapting,yao2018representation}. 
Other approaches attempted to tackle this issue by utilizing generative models, such as variational autoencoder (VAE) \cite{louizos2017causal,wu2021beta} and generative adversarial network (GAN) \cite{yoon2018ganite}, which utilize the favorable characteristics of generative models to generate either hidden unobserved variables, balanced latent variable, or uncertainties of counterfactual outcomes. These mentioned works have only shown encouraging results in estimating treatment effects from single-modality data. In practical scenarios, however, doctors routinely integrate both imaging and non-imaging data when making prognoses, and the interpretation of imaging data is substantially impacted by clinical information \cite{huang2020fusion}. In this regard, we consider two key ingredients. Firstly, the selection bias commonly exists in clinical scenarios, and the sysematic imbalance brought by this bias is amplified in high-dimensional data, as the higher number of covariates makes it more challenging to establish and verify overlap \cite{d2021overlap}. Therefore, it would be significant if we could map imbalanced high-dimensional data into a balanced low-dimensional representation. We thus seek to generate the distribution of low-dimensional prognostic score \cite{hansen2008prognostic}, which we will explain in Section 2 later. Secondly, motivated by the existing multi-modality VAE models \cite{kingma2013auto,lee2021variational,shi2019variational,wu2018multimodal}, we can fuse multi-modality distributions into a joint distribution with reasonable feasibility, which can be leveraged to construct a multi-modality model for prognosis.

In this paper, we propose a novel prognostic model that leverages both imaging and tabular data to achieve accurate treatment outcome prediction. This model is intended to be trained on observational data obtained from the non-randomized controlled trials. Specifically, to increase the reliability of the model, we employ a variational autoencoder model to generate a low-dimensional prognostic score that alleviates the problem of selection bias. Moreover, we introduce a variational distributions combination module that integrate information from imaging data and non-imaging clinical data to generate the aforementioned prognostic score. We evaluate our proposed model on a clinical dataset of intracerebral hemorrhage and demonstrate a significant improvement in treatment outcome prediction compared to existing treatment effect estimation techniques. 

\begin{figure}[!t]
    \centering
	\includegraphics[width =11.0cm]{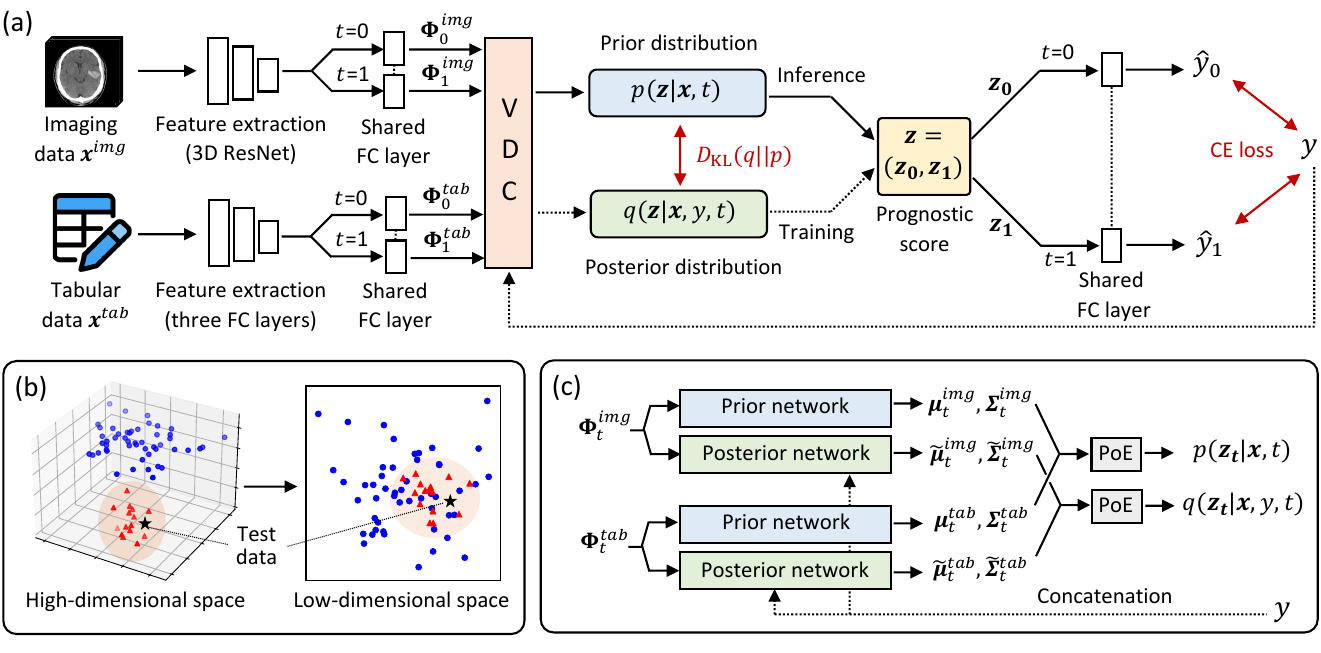}
	\caption{(a) Overview of proposed generative prognostic model. (b) Illustrative example: dimension reduction can increase the overlap between representation spaces of training samples with $T=0$ (blue points) and $T=1$ (red points), so that improves outcome prediction reliability for all treatment assignments (e.g., test data fall within the overlap). (c) Variational distribution combination (VDC) module; PoE denotes using the product-of-experts to generate joint distributions from different means and covariances.}
	\label{pipeline}
\vspace{-0.5cm}
\end{figure}

\section{Method}
\subsection{Formulation and Motivation}

We aim to predict the individualized treatment outcome based on a set of observations that include the actual treatment $T$, observed covariates $X$, and factual outcome $Y$. In this paper, we study the one-year functional outcome of patient who underwent either conservative treatment ($T=0$) or surgery ($T=1$). For each individual, let $t\in \{0, 1\}$ denote the treatment assignment, $\boldsymbol{x}=(\boldsymbol{x}{^{img}}$, $\boldsymbol{x}{^{tab}})$ represent the observed covariates comprising imaging data $\boldsymbol{x}{^{img}}$ and non-imaging tabular data $\boldsymbol{x}{^{tab}}$, and $y$ indicate the factual outcome. In this study, the treatment outcome was assessed using 1-year modified Rankin Scale (mRS) \cite{van1988interobserver}. Our objective is to estimate $\mathbb{E}[Y \mid X = \boldsymbol{x}, T=t]$.

The non-randomized controlled trials impacted by treatment preference can lead to selection bias, rendering the model unreliable due to potential encounters with unobserved scenarios during training. 
To address this issue, our model is inspired by the approach commonly used by doctors in clinical practice: using a combination of imaging data and non-imaging biomarkers to generate a prognostic score (e.g., GCS score) that predicts the likelihood of good or poor condition after treatment. 
In this study, a prognostic score is defined as any function $f_{T}(X)$ of $X$ and $T$ that Markov separates $Y$ and $X$, such that $Y \Vbar X | f_{T}(X)$. The insight is that a patient's health status can be effectively captured by a low-dimensional score $Z=f_{T}(X)$, which is a form of dimension reduction that is sufficient for causal inference and can naturally mitigate the problem brought by non-randomized controlled trials. This is because, as illustrated in Fig.~\ref{pipeline} (b), the difficulty of establishing and verifying overlap (between samples with $T=0$ and samples with $T=1$) increases in high-dimensional feature space compared to low-dimensional feature space \cite{d2021overlap}. We consider utilizing a VAE-based model for generating a prognostic score, due to two key ingredients: On the one hand, modeling score through a conditional distribution instead of a deterministic function offers greater flexibility \cite{wu2021beta}. On the other hand, VAE is a good model for dimension reduction, compared with vanilla encoder and other generative models. It has also been proved to be effective for treatment effect estimation.

\subsection{Generative Prognostic Model}

\textbf{Architecture.} As can be seen in Fig.~\ref{pipeline} (a), we first use two parallel networks to generate latent variables of imaging data and non-imaging tabular data, respectively. For the imaging data, we employ a 3D ResNet-34 \cite{he2016deep} as our feature extraction network and modify the final fully connected layers. We then generate the features conditioned on different treatments by concatenating the extracted features with their respective treatment assignments $t$ and forwarding them to a shared fully connected layer (FC layer) , yielding $\boldsymbol{\Phi}_{0}^{img}$ and $\boldsymbol{\Phi}_{1}^{img}$ respectively. This allows us to incorporate treatment assignment information and generate the prognostic score more effectively. For the non-imaging tabular data, we employ three blocks of a FC layer, followed by a Batch Normalization layer, and a ReLU activation function to generate the features $\boldsymbol{\Phi}_{0}^{tab}$ and $\boldsymbol{\Phi}_{1}^{tab}$. These features are then forwarded to a variational distribution combination (VDC) module, which we will describe in detail later. Through the VDC module, we can estimate the prior distribution $p(\boldsymbol{z}\mid \boldsymbol{x}, t)$ of the prognostic score $\boldsymbol{z}=(\boldsymbol{z_{0}},\boldsymbol{z_{1}})$. In addition, during the training phase, the true posterior distribution $q(\boldsymbol{z}\mid \boldsymbol{x}, y, t)$ can be approximated, which is additionally conditioned on $y$ and can help the model learn how to estimate an accurate prior distribution. The prognostic score $\boldsymbol{z_{0}},\boldsymbol{z_{1}}$ are then concatenated with treatment $t=1$ and $t=0$ respectively and are passed through a decoder consisting of a shared FC layer, to output the predicted potential outcomes with different treatment assignments, i.e, $\hat{y}_{0}$ and $\hat{y}_{1}$. Notably, during the inference phase, we use the prior distribution $p(\boldsymbol{z}\mid \boldsymbol{x}, t)$ to generate $\boldsymbol{z}$. In contrast, during the training phase, we use the posterior distribution $q(\boldsymbol{z}\mid \boldsymbol{x}, y, t)$ to generate $\boldsymbol{z}$ and predict the outcomes.
\\
\\
\textbf{Training scheme.} The evidence lower bound (ELBO) of our model is given by:
\begin{equation}
\begin{aligned}\label{eq1}
ELBO = \mathbb{E}_{\boldsymbol{z} \sim q} \log p(y \mid \boldsymbol{z}, t)-\beta D_{\mathrm{KL}}\left(q(\boldsymbol{z} \mid \boldsymbol{x}, y, t) \| p(\boldsymbol{z} \mid \boldsymbol{x}, t)\right),
\end{aligned}
\end{equation}
where $D_{\mathrm{KL}}\left(q(\boldsymbol{z} \mid \boldsymbol{x}, y, t) \| p(\boldsymbol{z} \mid \boldsymbol{x}, t)\right)$ is the Kullback-Leibler (KL) divergence between distributions $q(\boldsymbol{z} \mid \boldsymbol{x}, y, t)$ and $p(\boldsymbol{z} \mid \boldsymbol{x}, t)$, and $\beta$ is the weight balancing the terms in the ELBO. Note that the first term of Eq.~\ref{eq1} corresponds the classification error, which is minimized by the cross entropy loss. The second term of Eq.~\ref{eq1} uses KL divergence to encourage convergence of the prior distribution towards the posterior distribution. The training objective is to maximize the ELBO given the observational data, so that the model can be optimized.

\subsection{Variational Distributions Combination}
Once the features of imaging and non-imaging tabular data have been extracted, the primary challenge is to effectively integrate the multi-modal information. One approach that is immediately apparent is to train a single encoder network that takes all modalities as input, which can explicitly parameterize the joint distribution. Another commonly used method called Mixture-of-Experts (MoE) proposes to fuse the distributions from different modalities by weighting \cite{shi2019variational}. However, for this study, we generate the distribution of each modality separately and then use the Product-of-Experts (PoE) method to combine the two distributions into a single one \cite{lee2021variational,wu2018multimodal}. As can be seen in Fig.~\ref{pipeline} (c), assuming that we have generated distributions of two modalities with the means $\boldsymbol{\mu}^{img}_{t}$ and $ \boldsymbol{\mu}^{tab}_{t}$, and covariances $\boldsymbol{\Sigma}^{img}_{t}$ and $\boldsymbol{\Sigma}^{tab}_{t}$ respectively from the prior network, we can use PoE to generate the joint distributions $p(\boldsymbol{z}_{t}\mid \boldsymbol{x},t)=\mathcal{N} (\boldsymbol{\mu}_{t}, \boldsymbol{\Sigma}_{t})$ by:
\begin{equation}
\begin{aligned}\label{eq2}
\boldsymbol{\mu}_{t}=\left(\boldsymbol{\mu}_{t}^{pri}/\boldsymbol{\Sigma}_{t}^{pri}+\boldsymbol{\mu}_{t}^{img}/\boldsymbol{\Sigma}_{t}^{img}+\boldsymbol{\mu}_{t}^{tab}/\boldsymbol{\Sigma}_{t}^{tab}\right)\boldsymbol{\Sigma}_{t},
\end{aligned}
\end{equation}
\begin{equation}
\begin{aligned}\label{eq3}
\boldsymbol{\Sigma}_{t}=\left(1/\boldsymbol{\Sigma}_{t}^{pri}+1/\boldsymbol{\Sigma}_{t}^{img}+1/\boldsymbol{\Sigma}_{t}^{tab}\right)^{-1},
\end{aligned}
\end{equation}
where $\boldsymbol{\mu}_{t}^{pri}$ and $\boldsymbol{\Sigma}_{t}^{pri}$ are mean and covariance of universal prior expert, which is typically a spherical Gaussian ($\mathcal{N} (0, 1)$). For posterior distribution $q(\boldsymbol{z}_{t}\mid \boldsymbol{x}, y, t)$, we first additionally concatenate the features and $y$ together, and then generate the joint distribution by the same way. The PoE for generating joint distributions offers several advantages over the aforementioned approaches. Compared with the approaches that simply combing the features and then generating the joint distributions, PoE not only can effectively address the potential issue of prediction outcomes being overly influenced by the modality with a more abundant feature \cite{huang2020fusion}, but also is more flexible and has the potential to handle missing modalities. Compared to that using a mixture-of-experts, it can produce sharper distributions \cite{hinton2002training,lee2021variational}, which is desirable for our multi-modality data with complementary or unevenly distributed information.

\section{Experiment}
\subsection{Dataset and Experimental Setup}
\textbf{Datasets.} We utilized an in-house dataset of intracerebral hemorrhage cases obtained from the Hong Kong Hospital Authority. The dataset comprises 504 cases who underwent head CT scans and were diagnosed with ICH. Among them, 364 cases received conservative treatment, and 140 cases underwent surgery treatment. For each case, we collected both CT imaging and non-imaging clinical data. The non-imaging data have 17 clinical characteristics which have been proved to be potentially associated with the treatment outcome in clinical practice \cite{ji2013novel}, including gender, age, admission type, GCS, the history of smoking and drinking, hypertension, diabetes mellitus, hyperlipidemia, history of atrial fibrillation, coronary heart disease, history of stroke, pre-admission anticoagulation, pre-admission antiplatelet, pre-admission statin, small-vessel vascular disease and lower cholesterol. To address the selection bias resulting from the non-randomized controlled trials, we intentionally increased the imbalance of the dataset. Specifically, we selected 50 out of 68 cases who had IVH (another subtype of brain hemorrhage which can be infered from the CT image) and were treated conservatively, and 50 out of 61 cases who had a GCS score below 9 and underwent surgery. We used these samples for testing and reserved the remaining cases for training our model. As a result, the dataset is systematically imbalanced, which presents challenges for the model in producing reliable outcomes on test set. We also conducted additional experiments with different setting, as shown in the supplementary. A favorable outcome was defined as an mRS score of 0 to 3 (247 cases in total), while an unfavorable outcome was defined as an mRS score of 4 to 6 (257 cases in total) \cite{gregorio2018prognostic,hall2021identifying,stein2010spontaneous}. 

\textbf{Evaluation metrics.} We employed three evaluation metrics that are commonly used in treatment effect estimation and outcome prediction in our experiments, including the policy risk ($P_{ROL}$), the accuracy ($Acc$) and the area under the ROC curve ($AUC$). $P_{ROL}$ measures the average loss incurred when utilizing the treatment predicted by the treatment outcome estimator \cite{shalit2017estimating}, which is a lower-is-better metric.
Besides, we calculate $Acc_{0}$/$AUC_{0}$ for samples  factually treated with $T=0$ and $Acc_{1}$/$AUC_{1}$ for samples factually treated with $T=1$.

\textbf{Implementation details.} In preprocessing the imaging data, raw image intensity values were truncated to $[-20, 100]$, normalized to have zero mean and unit variance, and slices were uniformly resized to 224$\times$224 in the axial plane. We implemented our model using PyTorch and executed it on an NVIDIA A100 SXM4 card. For training, we used the Adam optimizer, a weight decay of $5\times10^{-3}$, and an initial learning rate of $5\times10^{-3}$. The training process lasted for a total of 2 hours, consisting of 1000 epochs with a batch size of 128. 
Our reported results are the average and standard deviation obtained from three independent runs.

\begin{table}[!t]
\centering
\caption{Comparison with state-of-the-art methods on the ICH dataset.}\label{tab1}
\scalebox{0.85}{
\begin{tabular}
% {cccccc}
{cp{2.2cm}<{\centering}p{2.2cm}<{\centering}p{2.2cm}<{\centering}p{2.2cm}<{\centering}cp{2.2cm}} 
\hline
\hline 
\multirow{2}{*}{Method} & \multicolumn{5}{c}{Evaluation metrics}  \\ \cmidrule(r){2-6} 
 & $R_{POL}$ $\downarrow$  & $AUC_{0}$ $\uparrow$ & $AUC_{1}$ $\uparrow $  & $Acc_{0}$ $\uparrow$  & $Acc_{1}$ $\uparrow$    \\ \hline
BNN \cite{johansson2016learning} & .581 $\pm$ .028 & .770 $\pm$ .032 & .724 $\pm$ .018 & .720 $\pm$ .035 & .720 $\pm$ .053 \\
% CFR-MMD \cite{shalit2017estimating} & .544 $\pm$ .062 & .773 $\pm$ .007 & .731 $\pm$ .011 & .733 $\pm$ .031 & .727 $\pm$ .023 \\ 
CFR-WASS \cite{shalit2017estimating} & .568 $\pm$ .020 & .792 $\pm$ .013 & .743 $\pm$ .013 & .747 $\pm$ .012 & .727 $\pm$ .046 \\ 
SITE \cite{yao2018representation} & .536 $\pm$ .040 & .789 $\pm$ .022 & .741 $\pm$ .022 & .733 $\pm$ .012 & .740 $\pm$ .020 \\
$\beta$-Intact-VAE \cite{wu2021beta} & .533 $\pm$ .044 & .797 $\pm$ .020  & .774 $\pm$ .012 & .773 $\pm$ .011 & .753 $\pm$ .023\\
DAFT \cite{polsterl2021combining} & .575 $\pm$ .023 & .782 $\pm$ .021 & .732 $\pm$ .024 & .740 $\pm$ .020 & .733 $\pm$ .031 \\ 
\hline
Ours & \textbf{.502 $\pm$ .023} & \textbf{.820 $\pm$ .019} & \textbf{.801 $\pm$ .018} & \textbf{.793 $\pm$ .012} & \textbf{.780 $\pm$ .040}\\ \hline \hline
\end{tabular}}
\end{table}

\subsection{Experiment results}
\textbf{Comparison with state-of-the-art methods}. We benchmarked our method against state-of-the-art approaches for treatment effect estimation, which are recognized as strong competitors in this field. These approaches include \textbf{BNN} \cite{johansson2016learning}, which is a representative work that balances the distribution of different treatment groups by discrepancy distance minimization, \textbf{CFR-WASS} \cite{shalit2017estimating}, which use separate heads to estimate the treatment effect and use Wasserstein Distance to balance the distribution, \textbf{SITE} \cite{yao2018representation}, which prioritizes hard samples to preserve local similarity while also balancing the distribution of data, and \textbf{$\beta$-Intact-VAE} \cite{wu2021beta}, which proposes to use a novel VAE model to generate low-dimensional representation conditioned on both covariates and treatment assignments to handle the selection bias problem. These methods were primarily developed for estimating treatment effects on single-modality data. To apply these methods to our multi-modality data, we utilized the same feature extraction architectures as our approach. The features extracted from these networks are concatenated. The fused features are then forwarded to the specific networks described in these papers. Moreover, we compared our method with \textbf{DAFT} \cite{polsterl2021combining}, which designs a block to suppress high-level concepts from 3D images while considering both image and tabular data, making it great for processing multi-modal data.

Table.~\ref{tab1} shows a comparison of results from different methods for estimating treatment outcomes in ICH. Our proposed method demonstrated a significant improvement in model performance compared to other methods, as evaluated using all five metrics. Compared with the classic methods based on balanced presentation learning, $\beta$-Intact-VAE seeks to generate balanced latent variable so that is more effective and can achieve more accurate performance in our experiment setting. However, this strategy lacks the ability to extract complementary information as it does not include a specially designed module for combining the distributions extracted from different modalities. Instead of simply concatenating features and generating a low-dimensional representation, our model utilizes the PoE technique to combine two low-dimensional distributions generated from two distinct modalities, which can effectively mitigate the risk of prediction outcomes being disproportionately influenced by the more feature-rich modality \cite{huang2020fusion}. For these
reasons, our proposed method improved performances by 3.1$\%$ on $R_{POL}$, 2.3$\%$/2.7$\%$ on $AUC_{0}$/$AUC_{1}$, and 2.0$\%$/2.7$\%$ on $Acc_{0}$/$Acc_{1}$, respectively. 

\begin{figure}[!t]
    \centering
	\includegraphics[width =12cm]{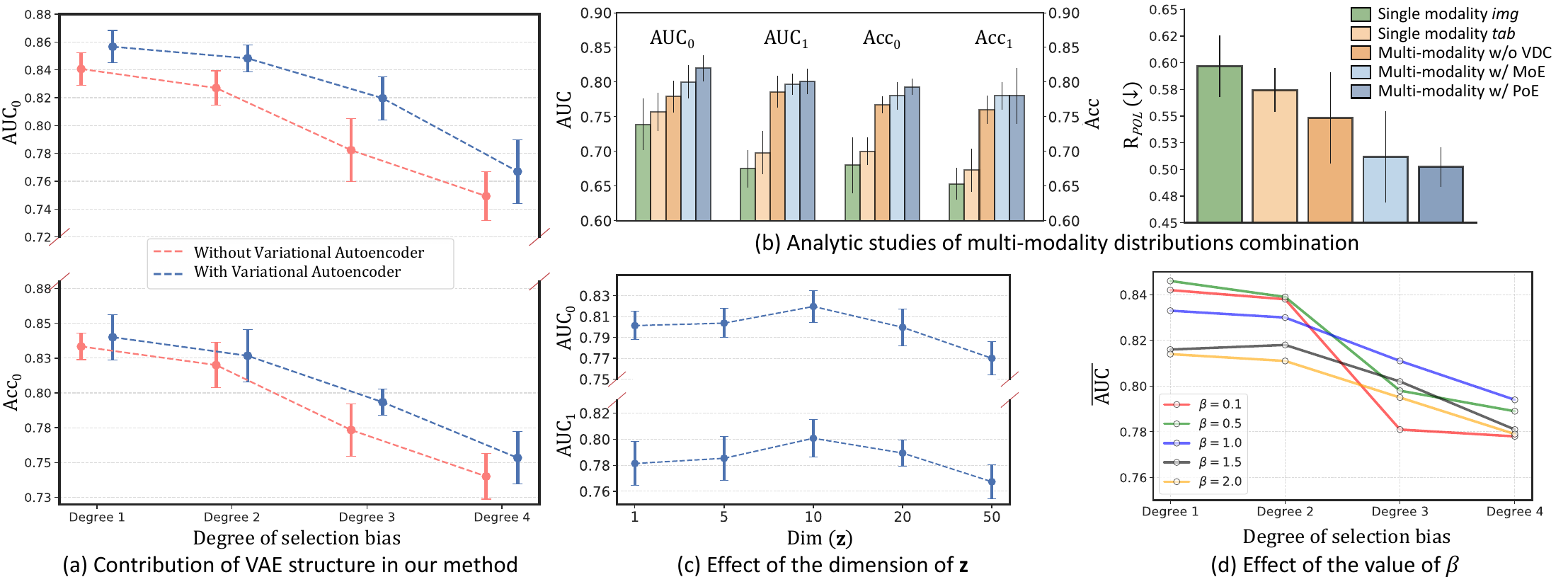}
	\caption{Ablation analysis of each component of our method.}
	\label{experiment}
\end{figure}

\textbf{Ablation analysis}. We then conducted comprehensive ablation studies. Initially, we studied the necessity of using a VAE structure for obtaining a low-dimensional prognostic score instead of a vanilla encoder. As shown in Fig.~\ref{experiment} (a), we systematically changed the degree of selection bias by varying the number of cases with IVH who underwent conservative treatment (68 in total) in the training set and test set. The ratios of training set/test set were: 68/0 (Degree 1), 48/20 (Degree 2), 18/50 (Degree 3) and 0/68 (Degree 4). When there are fewer cases in the training set, the selection bias increases, leading to a reduced ability of the model to predict such cases in the test set. The experiment showed that as the selection bias increased in the training set (ranging from Degree 1 to Degree 3), the difference between using a VAE structure and a vanilla encoder became more prominent. This is due to the VAE's effective dimension reduction. When there were no cases related to the outcome of interest (Degree 4) in the training set, further increases were stopped. This is expected since in such situations, there are no related cases that the model can learn from, thus rendering the advantages of dimension reduction ineffective.

Next, we studied the contributions of multi-modality distribution combination. As can be seen in Fig.~\ref{experiment} (b), despite using the proposed generative prognostic model, satisfactory performance cannot be achieved by simply using a single modality. Furthermore, compared to other commonly used approaches for integrating two generated distributions, such as simply combining the feature maps before generating the prognostic score (Multi-modality w/o VDC) and Mixture-of-Experts (Multi-modality w/ MoE), our proposed model (Multi-modality w/ PoE) achieved better performance. This highlights the effectiveness of distribution combination via PoE.  Additionally, in Fig.\ref{experiment} (c) and (d), we compared the performance of the model trained with different dimensions of generated prognostic score and the values of $\beta$ in Eq.\ref{eq1}. The dimension of the generated prognostic score is a trade-off between the degree of eliminating selection bias and the amount of accessible information. The results demonstrate that the optimal choice of dimension is 10. Moreover, note that $\beta$ controls the trade-off between outcome reconstruction and prognostic score recovery. Fig.\ref{experiment} (d) suggests $\beta$ should be 0.5 for low imbalance degree and 1.0 for high imbalance degree.

\section{Conclusion}

This paper introduces a novel generative prognostic model for predicting ICH treatment outcomes using imaging and non-imaging data. The model is designed to be trained on data collected from non-randomized controlled trials, addressing the imbalance problem with a VAE model and integrating multi-modality information using a variational distribution combination module. The model was evaluated on a large-scale dataset, confirming its effectiveness. 
\\
\\
{\textbf{Acknowledgement.} This work was supported in part by Shenzhen Portion of Shenzhen-Hong Kong Science and Technology Innovation Cooperation Zone under HZQB-KCZYB-20200089, in part by Hong Kong Innovation and Technology Commission Project No. ITS/238/21, in part by Hong Kong Research Grants Council Project No. T45-401/22-N, in part by Science, Technology and Innovation Commission of Shenzhen Municipality Project No. SGDX20220530111201008. We also thank the Hong Kong Hospital Authority Data Collaboration Laboratory (HADCL) for their support of this study.}

\footnotesize

\bibliographystyle{splncs04}
% \bibliography{paper2676}

\begin{thebibliography}{10}
\providecommand{\url}[1]{\texttt{#1}}
\providecommand{\urlprefix}{URL }
\providecommand{\doi}[1]{https://doi.org/#1}

\bibitem{bica2021real}
Bica, I., Alaa, A.M., Lambert, C., Van Der~Schaar, M.: From real-world patient
  data to individualized treatment effects using machine learning: current and
  future methods to address underlying challenges. Clinical Pharmacology \&
  Therapeutics  \textbf{109}(1),  87--100 (2021)

\bibitem{cheung2003use}
Cheung, R.T.F., Zou, L.Y.: Use of the original, modified, or new intracerebral
  hemorrhage score to predict mortality and morbidity after intracerebral
  hemorrhage. Stroke  \textbf{34}(7),  1717--1722 (2003)

\bibitem{ducruet2009complement}
Ducruet, A.F., Zacharia, B.E., Hickman, Z.L., Grobelny, B.T., Yeh, M.L.,
  Sosunov, S.A., Connolly~Jr, E.S.: The complement cascade as a therapeutic
  target in intracerebral hemorrhage. Experimental neurology  \textbf{219}(2),
  398--403 (2009)

\bibitem{d2021overlap}
D’Amour, A., Ding, P., Feller, A., Lei, L., Sekhon, J.: Overlap in
  observational studies with high-dimensional covariates. Journal of
  Econometrics  \textbf{221}(2),  644--654 (2021)

\bibitem{feigin2009worldwide}
Feigin, V.L., Lawes, C.M., Bennett, D.A., Barker-Collo, S.L., Parag, V.:
  Worldwide stroke incidence and early case fatality reported in 56
  population-based studies: a systematic review. The Lancet Neurology
  \textbf{8}(4),  355--369 (2009)

\bibitem{flaherty2006long}
Flaherty, M., Haverbusch, M., Sekar, P., Kissela, B., Kleindorfer, D., Moomaw,
  C., Sauerbeck, L., Schneider, A., Broderick, J., Woo, D.: Long-term mortality
  after intracerebral hemorrhage. Neurology  \textbf{66}(8),  1182--1186 (2006)

\bibitem{godoy2006predicting}
Godoy, D.A., Pinero, G., Di~Napoli, M.: Predicting mortality in spontaneous
  intracerebral hemorrhage: can modification to original score improve the
  prediction? Stroke  \textbf{37}(4),  1038--1044 (2006)

\bibitem{gregorio2018prognostic}
Greg{\'o}rio, T., Pipa, S., Cavaleiro, P., Atan{\'a}sio, G., Albuquerque, I.,
  Chaves, P.C., Azevedo, L.: Prognostic models for intracerebral hemorrhage:
  systematic review and meta-analysis. BMC Medical Research Methodology
  \textbf{18},  1--17 (2018)

\bibitem{gregson2019surgical}
Gregson, B.A., Mitchell, P., Mendelow, A.D.: Surgical decision making in brain
  hemorrhage: new analysis of the stich, stich ii, and stitch (trauma)
  randomized trials. Stroke  \textbf{50}(5),  1108--1115 (2019)

\bibitem{guo2022machine}
Guo, R., Zhang, R., Liu, R., Liu, Y., Li, H., Ma, L., He, M., You, C., Tian,
  R.: Machine learning-based approaches for prediction of patients’
  functional outcome and mortality after spontaneous intracerebral hemorrhage.
  Journal of Personalized Medicine  \textbf{12}(1), ~112 (2022)

\bibitem{hall2021identifying}
Hall, A.N., Weaver, B., Liotta, E., Maas, M.B., Faigle, R., Mroczek, D.K.,
  Naidech, A.M.: Identifying modifiable predictors of patient outcomes after
  intracerebral hemorrhage with machine learning. Neurocritical care
  \textbf{34},  73--84 (2021)

\bibitem{hansen2008prognostic}
Hansen, B.B.: The prognostic analogue of the propensity score. Biometrika
  \textbf{95}(2),  481--488 (2008)

\bibitem{he2016deep}
He, K., Zhang, X., Ren, S., Sun, J.: Deep residual learning for image
  recognition. In: CVPR. pp. 770--778 (2016)

\bibitem{hemphill2001score}
Hemphill~III, J.C., Bonovich, D.C., Besmertis, L., Manley, G.T., Johnston,
  S.C.: The ich score: a simple, reliable grading scale for intracerebral
  hemorrhage. Stroke  \textbf{32}(4),  891--897 (2001)

\bibitem{hemphill2015guidelines}
Hemphill~III, J.C., Greenberg, S.M., Anderson, C.S., Becker, K., Bendok, B.R.,
  Cushman, M., Fung, G.L., Goldstein, J.N., Macdonald, R.L., Mitchell, P.H.,
  et~al.: Guidelines for the management of spontaneous intracerebral
  hemorrhage: a guideline for healthcare professionals from the american heart
  association/american stroke association. Stroke  \textbf{46}(7),  2032--2060
  (2015)

\bibitem{hinton2002training}
Hinton, G.E.: Training products of experts by minimizing contrastive
  divergence. Neural computation  \textbf{14}(8),  1771--1800 (2002)

\bibitem{huang2020fusion}
Huang, S.C., Pareek, A., Seyyedi, S., Banerjee, I., Lungren, M.P.: Fusion of
  medical imaging and electronic health records using deep learning: a
  systematic review and implementation guidelines. NPJ digital medicine
  \textbf{3}(1), ~1--9 (2020)

\bibitem{ji2013novel}
Ji, R., Shen, H., Pan, Y., Wang, P., Liu, G., Wang, Y., Li, H., Zhao, X., Wang,
  Y.: A novel risk score to predict 1-year functional outcome after
  intracerebral hemorrhage and comparison with existing scores. Critical Care
  \textbf{17},  1--10 (2013)

\bibitem{johansson2016learning}
Johansson, F., Shalit, U., Sontag, D.: Learning representations for
  counterfactual inference. In: ICML. pp. 3020--3029. PMLR (2016)

\bibitem{kim2017spontaneous}
Kim, J.Y., Bae, H.J.: Spontaneous intracerebral hemorrhage: management. Journal
  of stroke  \textbf{19}(1), ~28 (2017)

\bibitem{kingma2013auto}
Kingma, D.P., Welling, M.: Auto-encoding variational bayes. In: ICLR (2014)

\bibitem{lee2021variational}
Lee, C., van~der Schaar, M.: A variational information bottleneck approach to
  multi-omics data integration. In: International Conference on Artificial
  Intelligence and Statistics. pp. 1513--1521. PMLR (2021)

\bibitem{louizos2017causal}
Louizos, C., Shalit, U., Mooij, J.M., Sontag, D., Zemel, R., Welling, M.:
  Causal effect inference with deep latent-variable models. NeurIPS
  \textbf{30} (2017)

\bibitem{peters2017elements}
Peters, J., Janzing, D., Sch{\"o}lkopf, B.: Elements of causal inference:
  foundations and learning algorithms. The MIT Press (2017)

\bibitem{polsterl2021combining}
P{\"o}lsterl, S., Wolf, T.N., Wachinger, C.: Combining 3d image and tabular
  data via the dynamic affine feature map transform. In: MICCAI. pp. 688--698.
  Springer (2021)

\bibitem{shalit2017estimating}
Shalit, U., Johansson, F.D., Sontag, D.: Estimating individual treatment
  effect: generalization bounds and algorithms. In: ICML. pp. 3076--3085. PMLR
  (2017)

\bibitem{shi2019adapting}
Shi, C., Blei, D., Veitch, V.: Adapting neural networks for the estimation of
  treatment effects. NeurIPS  \textbf{32} (2019)

\bibitem{shi2019variational}
Shi, Y., Paige, B., Torr, P., et~al.: Variational mixture-of-experts
  autoencoders for multi-modal deep generative models. NeurIPS  \textbf{32}
  (2019)

\bibitem{stein2010spontaneous}
Stein, M., Luecke, M., Preuss, M., Boeker, D.K., Joedicke, A., Oertel, M.F.:
  Spontaneous intracerebral hemorrhage with ventricular extension and the
  grading of obstructive hydrocephalus: the prediction of outcome of a special
  life-threatening entity. Neurosurgery  \textbf{67}(5),  1243--1252 (2010)

\bibitem{steiner2014european}
Steiner, T., Salman, R.A.S., Beer, R., Christensen, H., Cordonnier, C., Csiba,
  L., Forsting, M., Harnof, S., Klijn, C.J., Krieger, D., et~al.: European
  stroke organisation (eso) guidelines for the management of spontaneous
  intracerebral hemorrhage. International journal of stroke  \textbf{9}(7),
  840--855 (2014)

\bibitem{teasdale1974assessment}
Teasdale, G., Jennett, B.: Assessment of coma and impaired consciousness: a
  practical scale. The Lancet  \textbf{304}(7872),  81--84 (1974)

\bibitem{van1988interobserver}
Van~Swieten, J., Koudstaal, P., Visser, M., Schouten, H., Van~Gijn, J.:
  Interobserver agreement for the assessment of handicap in stroke patients.
  stroke  \textbf{19}(5),  604--607 (1988)

\bibitem{wu2018multimodal}
Wu, M., Goodman, N.: Multimodal generative models for scalable
  weakly-supervised learning. NeurIPS  \textbf{31} (2018)

\bibitem{wu2021beta}
Wu, P., Fukumizu, K.: $\backslash$beta-intact-vae: Identifying and estimating
  causal effects under limited overlap. ICLR  (2022)

\bibitem{yao2018representation}
Yao, L., Li, S., Li, Y., Huai, M., Gao, J., Zhang, A.: Representation learning
  for treatment effect estimation from observational data. NeurIPS  \textbf{31}
  (2018)

\bibitem{yoon2018ganite}
Yoon, J., Jordon, J., Van Der~Schaar, M.: Ganite: Estimation of individualized
  treatment effects using generative adversarial nets. In: ICLR (2018)

\end{thebibliography}

\end{document}